\documentclass{article}

\usepackage{arxiv}

\usepackage[utf8]{inputenc} 
\usepackage[T1]{fontenc}    
\usepackage{hyperref}       
\usepackage{url}            
\usepackage{booktabs}       
\usepackage{amsfonts}       
\usepackage{nicefrac}       
\usepackage{microtype}      
\usepackage{lipsum}		
\usepackage{graphicx}
\usepackage{natbib}
\usepackage{doi}
\usepackage{amsmath}
\usepackage{caption}
\usepackage{subcaption}
\usepackage{scrextend}

\title{Label Supervised LLaMA Finetuning\thanks{Preprint. Work in progress.}}

\date{} 					

\author{\textbf{Zongxi Li\hspace{.3mm}\textsuperscript{\rm 1}$^{\dagger}$, \hspace{.5mm}Xianming Li\hspace{.3mm}\textsuperscript{\rm 2}$^{\dagger}$, \hspace{.5mm}Yuzhang Liu\hspace{.3mm}\textsuperscript{\rm 3}, 
\hspace{.5mm}Haoran Xie\hspace{.3mm}\textsuperscript{\rm 4}, 
\hspace{.5mm}Jing Li\hspace{.3mm}\textsuperscript{\rm 2},}\\ 
\textbf{Fu-lee Wang\hspace{.3mm}\textsuperscript{\rm 1}, 
\hspace{.5mm}Qing Li\hspace{.3mm}\textsuperscript{\rm 2}, 
\hspace{.5mm}Xiaoqin Zhong\hspace{.3mm}\textsuperscript{\rm 3}} \\ \\
 $^1$ School of Science and Technology, Hong Kong Metropolitan University, Hong Kong SAR\\
$^2$ Department of Computing, Hong Kong Polytechnic University, Hong Kong SAR\\
$^3$ Shanghai 100me Internet Technology Co., Ltd., Shanghai, China\\
$^4$ Department of Computing and Decision Sciences, Lingnan University, Hong Kong SAR\\
$^{\dagger}$ Corresponding authors: \texttt{zoli@hkmu.edu.hk}, \texttt{xianming.li@connect.polyu.hk}
}

\renewcommand{\shorttitle}{LS-LLaMA}

\hypersetup{
pdftitle={Label Supervised LLaMA Finetuning},
pdfauthor={Zongxi Li, Xianming Li, Yuzhang Liu, Haoran Xie, Jing Li, Fu-lee Wang, Qing Li, Xiaoqin Zhong},
}

\begin{document}
\maketitle

\begin{abstract}

The recent success of Large Language Models (LLMs) has gained significant attention in both academia and industry. Substantial efforts have been made to enhance the zero- and few-shot generalization capabilities of open-source LLMs through finetuning. Currently, the prevailing approach is instruction-tuning, which trains LLMs to complete real-world tasks by generating responses guided by natural language instructions. It is worth noticing that such an approach may underperform in sequence and token classification tasks. Unlike text generation tasks, classification tasks have a limited label space, where precise label prediction is more appreciated than generating diverse and human-like responses. Prior research has unveiled that instruction-tuned LLMs cannot outperform BERT, prompting us to explore the potential of leveraging latent representations from LLMs for supervised label prediction. In this paper, we introduce a label-supervised adaptation for LLMs, which aims to finetuning the model with discriminant labels. We evaluate this approach with Label Supervised LLaMA (LS-LLaMA), based on LLaMA-2-7B, a relatively small-scale LLM, and can be finetuned on a single GeForce RTX4090 GPU. We extract latent representations from the final LLaMA layer and project them into the label space to compute the cross-entropy loss. The model is finetuned by Low-Rank Adaptation (LoRA) to minimize this loss. Remarkably, without intricate prompt engineering or external knowledge, LS-LLaMA substantially outperforms LLMs ten times its size in scale and demonstrates consistent improvements compared to robust baselines like BERT-Large and RoBERTa-Large in text classification. Moreover, by removing the causal mask from decoders, LS-unLLaMA achieves the state-of-the-art performance in named entity recognition (NER). Our work will shed light on a novel approach to adapting LLMs for various downstream tasks.

\end{abstract}


\section{Introduction}
Large Language Models (LLMs), such as GPT-3 \citep{brown2020language} and GPT-4 from OpenAI, LLaMA \citep{touvron2023llama, touvron2023llama2} from Meta, and PaLM \citep{chowdhery2022palm, anil2023palm} from Google, have demonstrated impressive language understanding and human-like response generation abilities. The large-scale pretraining and increased parameter size lead to phenomenal \emph{emergent abilities} \citep{wei2022emergent}, endowing LLMs with strong generalization capacity for unseen tasks, even in zero- and few-shot settings. Such capabilities of LLMs, unseen in smaller models, have significantly revolutionized methodologies across various natural language processing (NLP) tasks, ranging from label-supervised parameter optimization to prompt-based response generation. A key driving factor behind these transformations is the decoder-only architecture of these LLMs, which incorporates causal masks to prevent forward information exposure. Consequently, until now, the latent representations of LLMs have primarily been employed for predicting the next token in autoregressive text generation.

LLMs have been extensively evaluated in text classification and information extraction tasks, both in zero- and few-shot settings \citep{zhao2021calibrate, wei2023zero, li2023evaluating}, as well as with instruction tuning \cite{wang2023instructuie, lei2023instructerc}. Recent studies have revealed that zero-shot LLMs struggle to achieve satisfactory performance in these domains. For instance, GPT-3 (175B) achieves a classification accuracy of only 76\% on SST-2 and 43.9\% on AGNews \citep{zhao2021calibrate}, while GPT-3.5-Turbo (154B) attains an F1 score of 18.22\% for named entity recognition (NER) on OntoNotes \citep{wang2023instructuie}. Even the popular ChatGPT can only achieve F1 scores of 67.2\% and 51.1\% for NER on CoNNL2003 and OntoNotes, respectively \cite{li2023evaluating}. The above results fall short of state-of-the-art benchmarks. Although metric performance can be enhanced through careful prompt engineering and instruction-tuning techniques, these refined LLMs can hardly outperform discriminative encoder models such as BERT \citep{devlin-2019-bert} and RoBERTa \citep{liu2019roberta}.

We consider that these observations align with expectations, as LLMs and BERT possess different architectures and strengths that lead to divergent performance on specific tasks like text classification and NER. In classification tasks, labels often consist of highly concentrated words or phrases, and NER tasks involve symbolic tags, causing difficulties for LLMs in understanding the semantic meanings of the labels effectively. Although conversational LLMs can be finetuned on labeled datasets for these tasks, their generation-focused architecture may not capture task-specific patterns as efficiently as label-supervised BERT models, which have consistently demonstrated superior performance on a wide range of label prediction tasks. Furthermore, label spaces are considerably more restricted compared to the entire vocabulary, making the autoregressive generation less effective and efficient in label prediction, particularly in NER, where outputs are structured sequences of NER tokens.


Motivated by the efficacy of finetuning BERT and RoBERTa for classification tasks, we are the first to explore the feasibility of finetuning LLMs with label supervision to achieve effective task-specific adaptation. In this study, we employ the open-sourced LLaMA-2-7B model. We directly extract latent vector representations from the final LLaMA decoder layer, which was originally designed for autoregressive next-token prediction. These representations are then mapped into the label space through feed-forward layers, yielding logits that are used for discriminant label classification. We calculate the cross-entropy loss and employ Low-Rank Adaptation (LoRA) \citep{hu2021lora} to fine-tune the LLaMA model. Our preliminary results on multiclass benchmarks have been remarkably promising. We have observed significant improvements over both zero-shot and instruction-tuned LLaMA-2-7B models, as well as consistent enhancements compared to finetuned BERT. Extensive experiments demonstrate the effectiveness and robustness of our proposed task-specific LLaMA adaptation. We name such a configuration \textbf{L}abel \textbf{S}upervised \textbf{LLaMA} (LS-LLaMA). 

Before this work, open-sourced LLMs had yet been utilized for discriminant label classification as their latent representations were considered not suitable for language encoding, although Meta has provided the interface for sequence classification since LLaMA-1\footnote{\url{https://huggingface.co/docs/transformers/v4.33.2/model_doc/llama\#transformers.LlamaForSequenceClassification}} \citep{touvron2023llama}. Unlike the the encoder-based structure of Transformer \citep{vaswani2017attention} models, such as BERT and RoBERTa, conversational LLMs do not have an encoder structure \cite{brown2020language,touvron2023llama} and were trained primarily for language generation, focusing on predicting the next word in a sequence. The causal mask in the decoder blocks avoids forward information disclosure and also blocks bidirectional dependency extraction, leading to a myth that LLMs are not suitable for text encoding. Our research has has unveiled that the decoder output of LLMs contains a substantial amount of semantic meaning from the input text and can be effectively utilized as text representations for various classification tasks. However, the causal mask's presence causes fatal information loss at the token-level representation, leading to unsatisfactory results by LS-LLaMA in NER tasks. To overcome such a limitation, we propose an innovative solution by removing the causal mask from the decoders. After finetuning, the \textbf{L}abel \textbf{S}upervised \textbf{un}masked \textbf{LLaMA} (LS-unLLaMA) exhibits substantial improvements (up to 18\%) over LS-LLaMA in NER benchmarks i.e., CoNNL2003 and OntoNotes, and even outperforms LS-LLaMA on multiple text classification tasks. 

\begin{figure}[htbp]
    \centering
    \includegraphics[width=0.9\textwidth]{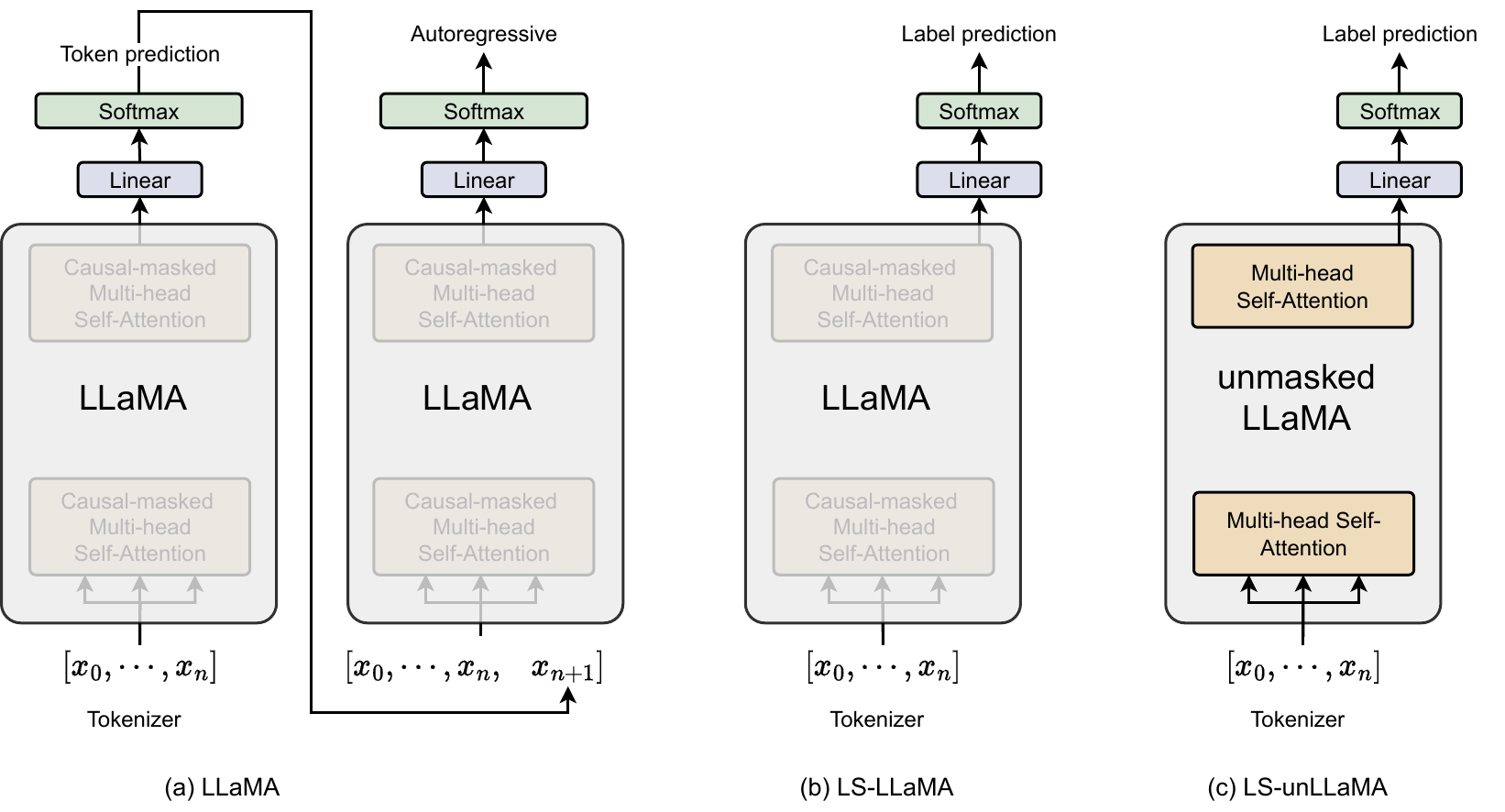}
    \caption{Comparison between conversational LLaMA and our proposed LS-LLaMA and LS-unLLaMA in sequence classification.}
    \label{fig:generic}
\end{figure}

This research aims to introduce a label-supervised adaptation configuration for LLMs. We investigate the feasibility of employing latent representations from LLMs for discriminant label prediction in text classification tasks. Our experimental results demonstrate that LLMs' latent representations can effectively serve as a text encoding method. Furthermore, we identify a limitation when using LLM's representations for token-level applications, attributed to the causal masks in the decoder structure, which restrict bidirectional information flow. To overcome this limitation, we remove the causal masks from LLaMA, resulting in state-of-the-art performance in NER tasks.

\section{Methodology}
This section presents the implementation of our proposed LS-LLaMA and LS-unLLaMA. Figure \ref{fig:generic} depicts how they are different from the conventional autoregressive settings.

\subsection{Label-supervised LLaMA}
We tokenize the input sequence $S$ with the default \verb+AutoTokenizer+ following the automatic operation of the transformers\footnote{\url{https://github.com/huggingface/transformers}} library. The tokens $T$ were fed into pretrained models through \verb+LlamaForSequenceClassification+ to extract the latent representation $H$ from LLaMA for sequence classification,
\begin{equation}
\begin{split}
T & = \mathrm{Tokenizer}(S) \\
H_{seq} & = \mathrm{LlamaForSeqClf}(T).
\end{split}
\end{equation}
The pooling operation is applied to the latent representation to obtain the vector representation $h$ for sequence classification. \verb+LlamaForSequenceClassification+'s default pooling operation takes out the last token vector from the final representation, as this is the only token that encodes all the historical information due to single-direction information flow caused by the causal masks. After passing through fully connected layers and a softmax layer, vector representation $h$ is mapped to the label space. Cross-entropy loss is calculated based on the output logits and the ground-truth label. 

LLaMA did not provide the interface for token classification. Therefore, we modify the LLaMA model to obtain all the token representations with \verb+LlamaForTokenClassification+\footnote{Code for LS-LLaMA is available on GitHub: \url{https://github.com/4AI/LS-LLaMA}}, $H_{tkn} = \mathrm{LlamaForTokenClf}(T)$, for token classification tasks.

We apply LoRA \citep{hu2021lora} to finetune the LLaMA model to maximize the probability of the correct label.

\subsection{Label-supervised unmasked LLaMA}
\begin{figure}
    \centering
    \includegraphics[width=0.9\textwidth]{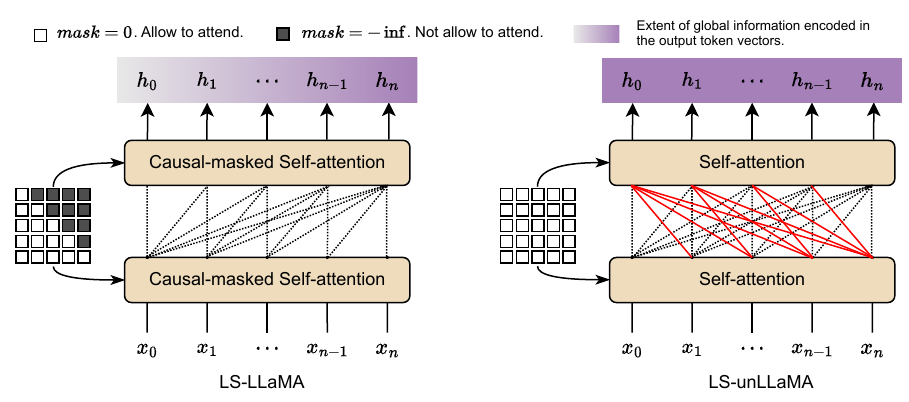}
    \caption{An illustration of the effects caused by different masking methods in LS-LLaMA and LS-unLLaMA.}
    \label{fig:llama_unllama}
\end{figure}

The causal masks $CM$, as shown in Equation \ref{eq:cm}, 
\begin{equation}
CM=\begin{bmatrix}
0 & -\inf & -\inf & \cdots & -\inf & -\inf & -\inf\\
0 & 0 & -\inf & \cdots & -\inf &-\inf & -\inf \\
0 & 0 & 0 & \cdots & -\inf &-\inf & -\inf \\
\cdots & \cdots & \cdots & \cdots & \cdots &\cdots &\cdots \\
\cdots & \cdots & \cdots & \cdots & \cdots &\cdots &\cdots \\
0 & 0 & 0 &\cdots & 0 &-\inf & -\inf\\
0 & 0 & 0 &\cdots &0 & 0&-\inf\\
0 & 0 & 0 & \cdots & 0 & 0 & 0
\end{bmatrix},
\label{eq:cm}
\end{equation}
in decoder blocks prevent information leaking, as the decoder is only allowed to attend to earlier positions in text generation. Bidirectional dependency extraction of the self-attention layer is reduced to single-direction, leading to critical information loss at the token level. Our empirical studies show that using token representations learned with causal masks significantly underperforms in token classification tasks. To address such an issue, we remove the causal masks from \verb+LlamaForTokenClassification+\footnote{Note that this operation is different from the padding mask and requires modification on LLaMA's source code.} and extract the latent representations $H'$ for token classification, as illustrated in Figure \ref{fig:llama_unllama},
\begin{equation}
    H'_{tkn} = \mathrm{unmasked\_LlmaForTokenClf}(T).
\end{equation}
The essential bidirectional information is expected to be replenished in token representations during finetuning as all the tokens can attend to each other. 

We believe the essential global dependency is also helpful in the sequence classification, and hence, remove the causal masks in \verb+LlamaForSequenceClassification+. With bidirectional self-attention resumed, we have more choices in pooling. We have tested three pooling methods, i.e., \verb|max|, \verb|average|, and \verb|last|, and the experiments show that max-over-time pooling yields better performance than average pooling and last-token pooling in classification tasks without causal masks.

\section{Experiment}

\subsection{Tasks and evaluation metrics}
We have conducted extensive experiments on text classification and NER tasks against zero- and few-shot LLMs, instruction-following LLMs, and discriminant baselines to validate the effectiveness of the proposed label-supervised adaptation method based on LLaMA-2-7B.

\subsubsection{Multiclass text classification}
Experiments were conducted on four English datasets from multiple domains, i.e., SST2 and SST5\footnote{\url{https://huggingface.co/datasets/SetFit/sst5}} (general sentiment analysis), AGNews (news categorization), and Twitter Financial News\footnote{\url{https://huggingface.co/datasets/zeroshot/twitter-financial-news-sentiment}} (short as ``Twitter-Fin'', financial sentiment analysis). Table \ref{tab:multiclass_data} contains the statistics and label classes of the datasets.

Moreover, we experimented on the German, English, Spanish, and Chinese subsets of Multilingual Amazon Reviews Corpus\footnote{\url{https://huggingface.co/datasets/amazon_reviews_multi}} \citep{marc_reviews} to test LS-LLaMA and LS-unLLaMA under the multilingual setting. The datasets contain Amazon product reviews for 31 product categories. The task is to predict the product category given the review content. For each language, the subset contains 200,000, 5,000, and 5,000 data samples in the training, development, and test sets respectively. We report the accuracy of text classification tasks.

\begin{table}[htbp]
	\caption{Dataset descriptions for multiclass text classification.}
	\centering
	\begin{tabular}{lrrl}
		\toprule
		Dataset & Train & Test & Classes \\
            \midrule
            SST2 & 67,300 & 872$^{*}$ & ``Positive'', ``Negative''\\
            SST5 & 8,540 & 2,210 &  ``Very negative'', ``Negative'', ``Neutral'', ``Positive'', ``Very positive''\\
            AGNews & 120,000 & 7,600 & ``World'', ``Sports'', ``Business', ``Sci/Tech''\\
            Twitter Fin & 9,938 & 2,486 &  ``Bearish'', ``Bullish'', ``Neutral'' \\
		\bottomrule
            \multicolumn{4}{l}{$^*$ Testing on the validation set for all baselines}
	\end{tabular}
	\label{tab:multiclass_data}
\end{table}

\subsubsection{Named entity recognition}
For the NER task, we experiment on OntoNotes V5.0\footnote{\url{https://catalog.ldc.upenn.edu/LDC2013T19}} and CoNLL2003 \citep{tjong-kim-2003-introduction}, which are widely-adopted datasets for information retrieval. OntoNotes V5.0 has 59,924 training samples and 8,262 testing samples, and CoNLL2003 contains 14,987 training samples and 3,684 testing samples. We report the F1 scores on the NER tasks. 

\subsection{Implementation details}
LLaMA-2-chat under the zero-shot setting was tested on the classification tasks. LLaMA-2-chat-7B and LLaMA-2-chat-13B were deployed on Nvidia GeForce RTX4090 and A100 GPUs, respectively.

The main experiments of label-supervised adaptation are based on LLaMA-2-7B. These experiments were run on a single Nvidia GeForce RTX4090 GPU. We adopt the standard parameter-efficient fine-tuning method LoRA to finetune the model. We configured the LoRA settings \textit{lora\_rank}, \textit{lora\_alpha}, and \textit{lora\_dropout} to $12$, $32$, and $0.1$, respectively. We set the batch size to 8 and initial learning rate to $8e-5$ using grid search. For the sequence classification, we use the \textit{last token} pooling on LS-LLaMA and the \textit{max} pooling on LS-unLLaMA according to the ablation study on the pooling method.

We apply instruction-tuning based on LLaMA-2-7B on text classification tasks. The prompt is set as follows:
\begin{addmargin}[2em]{2em}
    \verb|prompt_input =| ``You are a classification model. Based on the given article, you need to predict the most relevant category label from \verb|{all_labels}|. One article has only one label. {\textbackslash n} \#\#\# Input article: \verb|{input_text}| {\textbackslash n} \#\#\# Output: ''
\end{addmargin}
The instruction-following model is finetuned for 50k steps on large datasets (training samples exceed 50,000) and 10k steps on small datasets (training samples less than 50,000) with a batch size of 16 and an initial learning rate of $2e-4$.

\subsection{Experimental results}
\label{sec:results}

\begin{table}[htbp]
	\caption{Accuracy (\%) in multiclass text classification.}
	\centering
	\begin{tabular}{lcccc}
		\toprule
		Models & \textbf{SST2} & \textbf{AGNews} & \textbf{Twitter-Fin} & \textbf{SST5} \\
		\midrule
            \multicolumn{5}{l}{\underline{\textit{Zero- and few-shot}}}\\
            LLaMA-2-7B (zero-shot) & $76.26$ & $37.39$ & $23.40$ & $39.05$\\
            LLaMA-2-13B (zero-shot) & $69.90$ & $59.40$ & $38.74$ & $37.01$ \\
            GPT-3 175B (zero-shot) & $54.3^{\dagger}$ & $43.9^{\dagger}$ & $-$& $-$ \\
            GPT-3 175B (few-shot) & $93.4^{\dagger}$ & $84.3^{\dagger}$ & $-$& $-$ \\
            \midrule
            \multicolumn{5}{l}{\underline{\textit{Instruction-tuning}}}\\
            LLaMA-2-7B (instruction-tuning) & $91.97$ & $52.40$ & $68.72$ & $43.35$\\

		\midrule
            \multicolumn{5}{l}{\underline{\textit{Discriminant baselines}}}\\
            BERT-Base & $92.78$ & $94.51$ & $88.19$  & $55.07$\\
            BERT-Large & $92.86$ &  $94.45$ & $88.74$ & $55.79$ \\
            RoBERTa-Base & $94.61$ & $94.70$ & $90.32$  & $58.46$\\
            RoBERTa-Large & $96.10$ & $94.78$ & $90.95$ & $59.64$ \\
            AGN \citep{li2021merging} & $93.27$ & $93.82$ & $-$ & $55.72$ \\
            \midrule
            \multicolumn{5}{l}{\underline{\textit{Ours}}}\\
            LS-LLaMA-2-7B & $96.67$ & $95.38$ & $\mathbf{91.87}$ & $\mathbf{62.31}$\\
            LS-LLaMA-2-13B & $96.90$ & $95.66$ & $91.20$&  $62.17$  \\
            LS-unLLaMA-2-7B & $\mathbf{97.36}$ & $\mathbf{95.68}$ & $91.54$ & $60.50$\\
            LS-unLLaMA-2-13B & $92.77$ & $95.44$ & $87.94$ & $52.99$ \\
		\bottomrule
            \multicolumn{3}{l}{$^{\dagger}$ results reported in \cite{zhao2021calibrate}}
	\end{tabular}
	\label{tab:multiclass}
\end{table}

\subsubsection{Multiclass text classification}
\label{sec:multiclass_result}
We present the results of multiclass text classification experiments in Table \ref{tab:multiclass}, offering a comprehensive analysis of LLMs under different adaptation settings and discriminant models. Notably, LLMs such as LLaMA-2-7B/13B and GPT-3-175B exhibit suboptimal performance when subjected to zero- and few-shot settings. While instruction-tuning brings about notable enhancements on LLaMA-2-7B, it is apparent that the achieved results are not comparable to the discriminative capabilities demonstrated by BERT and RoBERTa. This observation highlights the challenges faced by conversational LLMs when predicting from a predefined label set.

Our proposed label-supervised adaptation method showcases substantial improvements over instruction fine-tuning. Notably, the magnitude of these enhancements is more outstanding in datasets where labels are more domain-specific and demand a deeper understanding of commonsense knowledge. For instance, in SST2, which features commonly-used binary sentiment labels (``Positive'' and ``Negative''), we observe an absolute improvement of 5.39\% (equivalent to a relative improvement of 5.86\%). In contrast, in SST5 with fine-grained sentiment labels like ``Neutral'', ``Very positive'', and ``Very negative'', our approach demonstrates an 18.96\% absolute improvement (equivalent to a relative improvement of 43.74\%). When applied to the Twitter-Fin dataset, which comprises domain-specific financial labels (``Bullish'' and ``Bearish''), our method produces a 23.15\% absolute improvement (equivalent to a relative improvement of 33.69\%). In the AGNews dataset, labeled with domain-specific news categories, we achieve a 43.28\% absolute improvement (equivalent to an astonishing relative improvement of 82.60\%). Remarkably, such improvements do not require sophisticated prompt engineering and external knowledge.

Furthermore, it is noteworthy that both LS-LLaMA and LS-unLLaMA consistently demonstrate significant improvements when compared to robust discriminative baselines such as BERT-Large and RoBERTa-Large. Our approach achieves improvements of 1.26\%, 0.90\%, 0.92\%, and 2.67\% when compared against RoBERTa-Large on SST2, AGNews, Twitter-Fin, and SST5, respectively. Interestingly, we also observe variations in performance between LS-LLaMA and LS-unLLaMA. Specifically, LS-unLLaMA outperforms LS-LLaMA on SST2 and AGNews, but exhibits inferior performance on Twitter-Fin and SST5. A potential explanation for these differences can be drawn from the dataset sizes, as indicated in Table \ref{tab:multiclass_data}. SST2 and AGNews are notably larger datasets compared to Twitter-Fin and SST5, a distinction that may provide insights. Since we removed causal masks from LS-unLLaMA, the model needs more training samples to reconstruct the parameters that were previously concealed during pretraining. This process is not strictly needed in LS-LLaMA as it can directly employ the last token for classification task. Therefore, we conclude that LS-LLaMA can quickly adapt on small-scale datasets, and LS-unLLaMA can achieve even better results with ample training samples.

\begin{table}[htbp]
	\caption{Accuracy (\%) in multiclass text classification on Multilingual Amazon Reviews Corpus.}
	\centering
	\begin{tabular}{lccccc}
		\toprule
		Models & \textbf{DE} & \textbf{EN} & \textbf{ES} & \textbf{ZH} & \textbf{Avg.}\\
		\midrule
            \multicolumn{5}{l}{\underline{\textit{Zero-shot}}}\\
            LLaMA-2-7B (zero-shot) & $9.23$ & $12.13$ & $7.43$ & $19.65$ & $12.11$\\
            \midrule
            \multicolumn{5}{l}{\underline{\textit{Discriminant baselines}}}\\
            BERT-Base-Multilingual & $53.54$ & $53.42$ & $46.22$  & $67.35$ & $55.13$ \\
            RoBERTa-Base-Multilingual & $52.61$ & $52.56$ & $44.32$  & $66.98$ & $54.12$\\
            RoBERTa-Large-Multilingual & $52.76$ & $53.64$ & $44.76$ & $66.90$ & $54.52$ \\
            \midrule
            \multicolumn{5}{l}{\underline{\textit{Ours}}}\\
            LS-LLaMA & $56.80$ & $58.82$ & $49.28$ & $68.72$ & $58.41$\\
            LS-unLLaMA & $\mathbf{56.90}$ & $\mathbf{60.20}$ & $\mathbf{49.68}$ & $\mathbf{69.70}$ & $\mathbf{59.21}$\\
		\bottomrule
	\end{tabular}
	\label{tab:multilingual}
\end{table}

\subsubsection{Multilingual multiclass text classification}
According to \cite{touvron2023llama2}, LLaMA-2's pretraining data primarily consists of English text, accounting for a substantial 89.70\%, with non-English languages representing only a minority within the pretraining dataset—specifically, 0.17\% German, 0.13\% Chinese, and 0.13\% Spanish. However, ``[a] training corpus with a majority in English means that the model may not be suitable for use in other languages.'' Consequently, we conducted a series of tests to evaluate LLaMA's text classification capabilities under the zero-shot setting and with our proposed label-supervised adaptation method within multilingual environments.
In particular, the adopted Amazon Reviews Corpus encompasses 31 product category labels, making the classification more difficult for both LLMs and discriminant baselines. Zero-shot LLaMA-2-7B presents low accuracy when asked to select one label from 31 candidates. With label-supervised finetuning,  the model's performance gains improvements ranging from four to sixfold. In particular, compared with the multilingual version of BERT and RoBERTa, our approaches can still gain remarkable improvements of at least 3.36\%, 6.56\%, 3.46\%, and 2.35\% on the German, English, Spanish, and Chinese subsets, respectively, underlining the exceptional multilingual learning capacity of LLaMA. These results underscore the superior efficacy of the proposed label-supervised adaptation method when applied to a range of text classification tasks. Furthermore, LS-unLLaMA outperforms LS-LLaMA, suggesting the benefit brought by the removal of the causal mask.

\begin{table}[htbp]
	\caption{F1 score (\%) in NER.}
	\centering
	\begin{tabular}{lcc}
		\toprule
		Models & \textbf{CoNNL2003} & \textbf{OntoNotes V5}\\
		\midrule
            \multicolumn{3}{l}{\underline{\textit{Zero- and few-shot}}}\\
            LLaMA-2-7B (zero-shot) & $1.35$ &$1.20$ \\
            ChatGPT & $67.20^{\dagger}$ & $51.10^{\dagger}$\\
            GPT-3.5-Turbo & $-$ & $18.22^{\ddagger}$\\
            \midrule
            \multicolumn{3}{l}{\underline{\textit{Instruction-tuning}}}\\
            InstructUIE \citep{wang2023instructuie} & $92.94$ & $90.19$ \\

		\midrule
            \multicolumn{3}{l}{\underline{\textit{Discriminant baselines}}}\\
            BERT-Base & $92.40^{\mathsection}$ & $88.88^\star$ \\
            BERT-Large & $92.80^{\mathsection}$ &  $89.27$  \\
            RoBERTa-Base & $92.13$ & $91.55$ \\
            RoBERTa-Large & $92.59$ & $91.72$\\
            RAN \citep{li-etal-2023-recurrent} & $92.68$ & $89.38$ \\
            \midrule
            \multicolumn{3}{l}{\underline{\textit{Ours}}}\\
            LS-LLaMA-2-7B & $74.76$ & $77.41$ \\
            LS-LLaMA-2-13B & $74.12$ &  $77.73$\\
            LS-unLLaMA-2-7B & $\mathbf{93.19}$ & $\mathbf{92.10}$\\
            LS-unLLaMA-2-13B & $91.46$ & $91.07$ \\
		\bottomrule
            \multicolumn{3}{l}{$^{\dagger}$ results reported in \cite{li2023evaluating}}\\
            \multicolumn{3}{l}{$^{\ddagger}$ results reported in \cite{wang2023instructuie}}\\
            \multicolumn{3}{l}{$^{\mathsection}$ results reported in \cite{devlin-2019-bert}}\\
            \multicolumn{3}{l}{$^{\star}$ results reported in \cite{li-etal-2023-recurrent}}
	\end{tabular}
	\label{tab:ner}
\end{table}

\subsubsection{Named entity recognition}
Results on NER tasks were presented in Table \ref{tab:ner}. The highlight of the results lies in the performance margin between LS-LLaMA and LS-unLLaMA. In text classification tasks, these two variants can produce comparable results. However, LS-unLLaMA achieves 18.43\% and 14.69\% higher F1 scores than LS-LLaMA on CoNNL2003 and OntoNotes, respectively. 

The results pertaining to NER tasks are outlined in Table \ref{tab:ner}. LLaMA-2-7B under zero-shot setting can only produce single-digit result of F1 score, showing that LLaMA does not have suitable knowledge for addressing NER tasks. Our approach presents higher performance than BERT-Large and RoBERTa-Large. What deserves special attention is the distinctive performance margin observed between LS-LLaMA and LS-unLLaMA. In the realm of text classification tasks, these two variants tend to yield comparable results. However, in NER tasks, LS-unLLaMA exhibits much more reliable performance. Specifically, LS-unLLaMA achieves F1 scores that are 18.43\% and 14.69\% higher than those achieved by LS-LLaMA on the CoNLL2003 and OntoNotes datasets, respectively. This performance gap indicates the pivotal of global dependencies, encoded within token representations and facilitated by bidirectional self-attention mechanisms, in effectively addressing NER challenges. Such features were absent in LS-LLaMA and other LLMs.

\subsection{Ablation study}
\subsubsection{Pooling method}

We conducted a comprehensive study of various pooling methods, including \verb|max|, \verb|average|, and \verb|last|, on the SST2 and SST5 datasets to determine their impact on model performance. The findings, as shown in Table \ref{tab:pool}, highlight distinct trends. Notably, when employing causal masks, LS-LLaMA achieves the best performance when adopting the \verb|last| pooling method. Conversely, LS-unLLaMA without causal masks demonstrates optimal capabilities when utilizing the \verb|max| pooling strategy.

The reason of such a difference attributes to the presence or absence of causal masks. In LS-LLaMA, causal masks enforces a single-direction information flow, allowing only the last token to attend to all the previos tokens. Consequently, the semantic meaning of the entire sentence aggregates at the last token. In contrast, the tokens within LS-unLLaMA, free from the constraints of causal masks, possess the capacity to attend to every other token in a bidirectional manner. The expanded connectivity facilitates the utilization of max-over-time pooling, which facilitates extracting global dependencies within the sentence.

\begin{table}
	\caption{Comparison between different pooling strategies}
	\centering
	\begin{tabular}{lccccc}
		\toprule
		& \multicolumn{2}{c}{SST2} & & \multicolumn{2}{c}{SST5}                   \\
		\cmidrule(r){2-3}
            \cmidrule(r){5-6}
		 & LS-LLaMA & LS-unLLaMA & &LS-LLaMA & LS-unLLaMA \\
        & w/ CM & w/o CM & & w/ CM & w/o CM\\
		\midrule
        \verb|max| &  $95.53$ & $\mathbf{97.36}$ & & $58.79$  & $\mathbf{60.50}$ \\
        \verb|average| & $55.92$  & $95.96$  & & $53.80$ & $59.41$ \\
        \verb|last| & $\mathbf{96.67}$ & $95.76$ & & $\mathbf{62.31}$ &  $46.74$ \\

		\bottomrule
\multicolumn{4}{l}{CM stands for causal mask.}
	\end{tabular}
	\label{tab:pool}
\end{table}

\subsubsection{Training process}
\begin{figure}[htbp]
     \centering
     \begin{subfigure}[b]{0.45\textwidth}
         \centering
         \includegraphics[width=\textwidth]{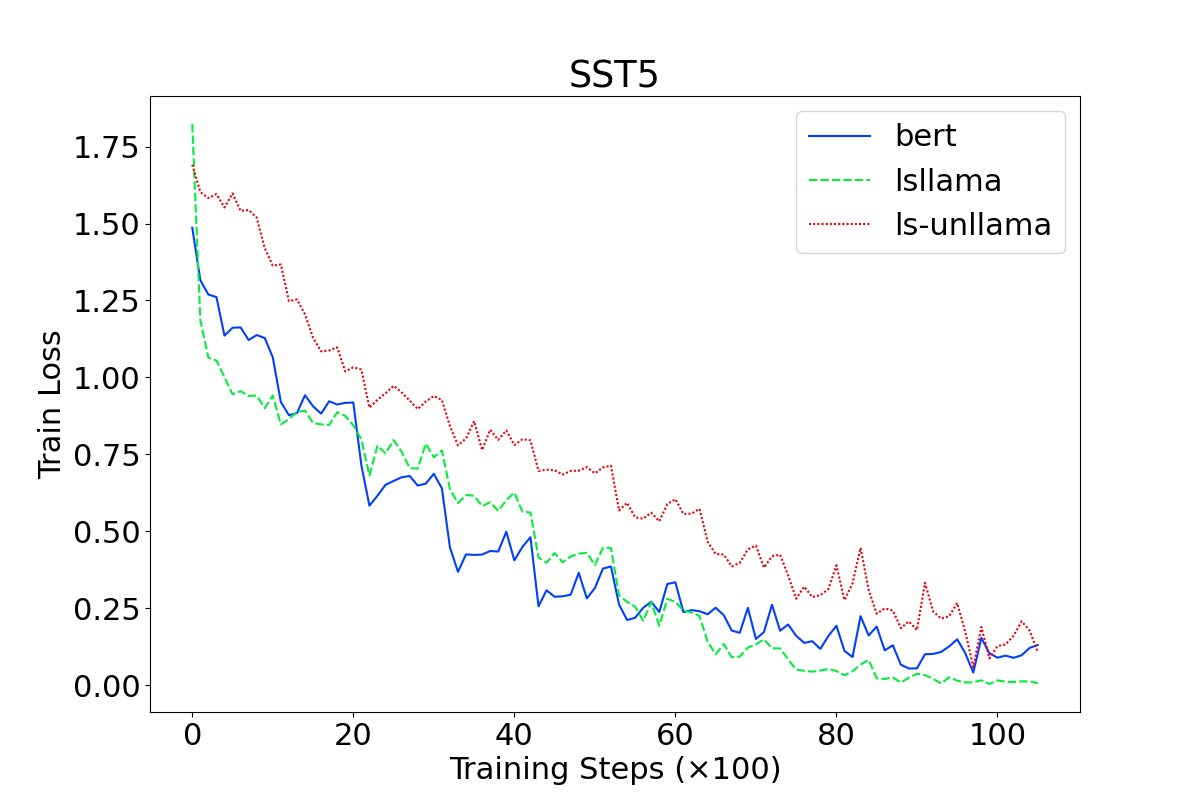}
     \end{subfigure}
     \begin{subfigure}[b]{0.45\textwidth}
         \centering
         \includegraphics[width=\textwidth]{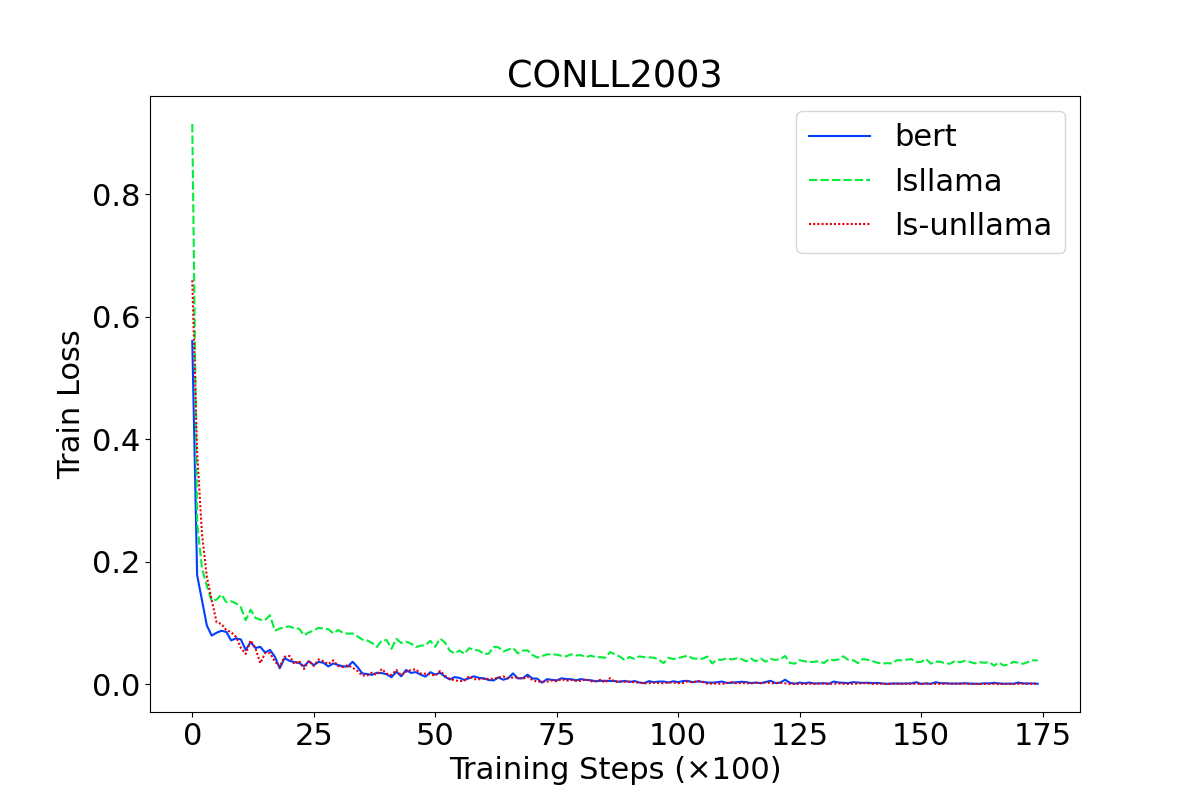}
     \end{subfigure}
     
        \caption{Training loss of BERT, LS-LLaMA, and LS-unLLaMA in 10 training epochs.}
        \label{fig:sst5_conll_train_loss}
\end{figure}

\begin{figure}[htbp]
     \centering
     \begin{subfigure}[b]{0.45\textwidth}
         \centering
         \includegraphics[width=\textwidth]{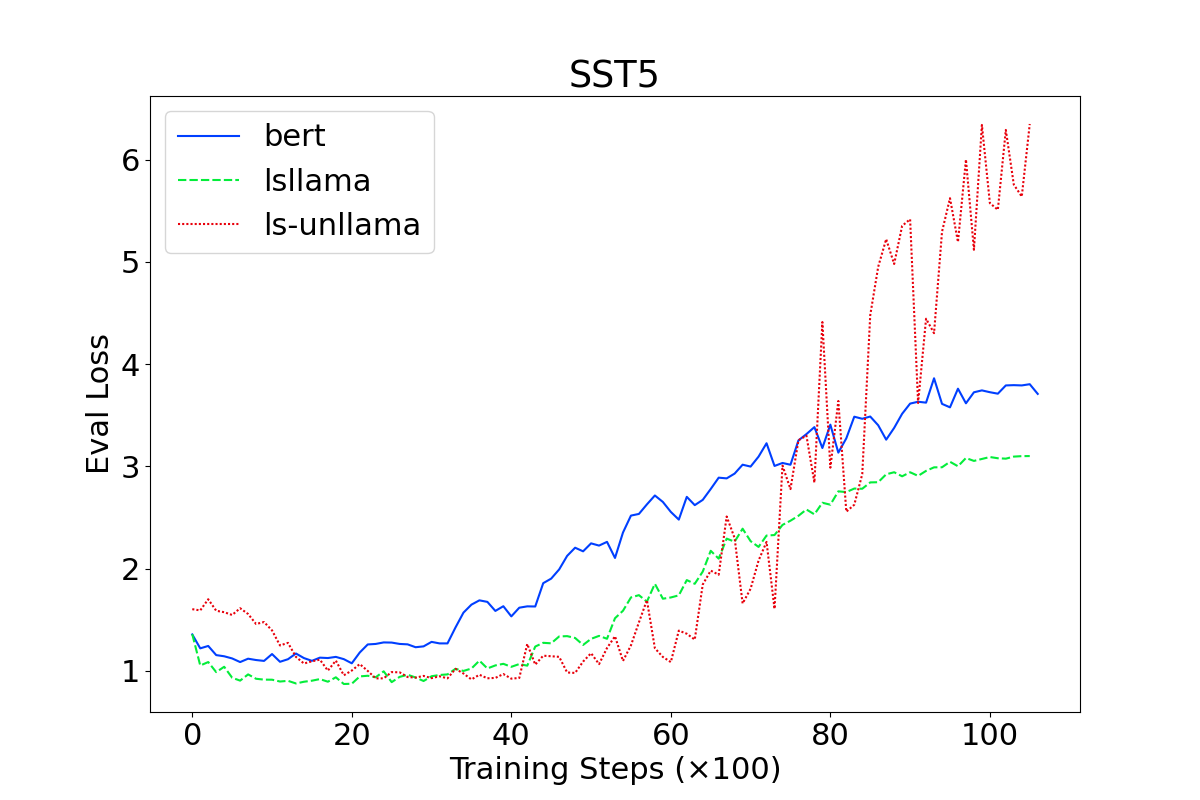}
     \end{subfigure}
     \begin{subfigure}[b]{0.45\textwidth}
         \centering
         \includegraphics[width=\textwidth]{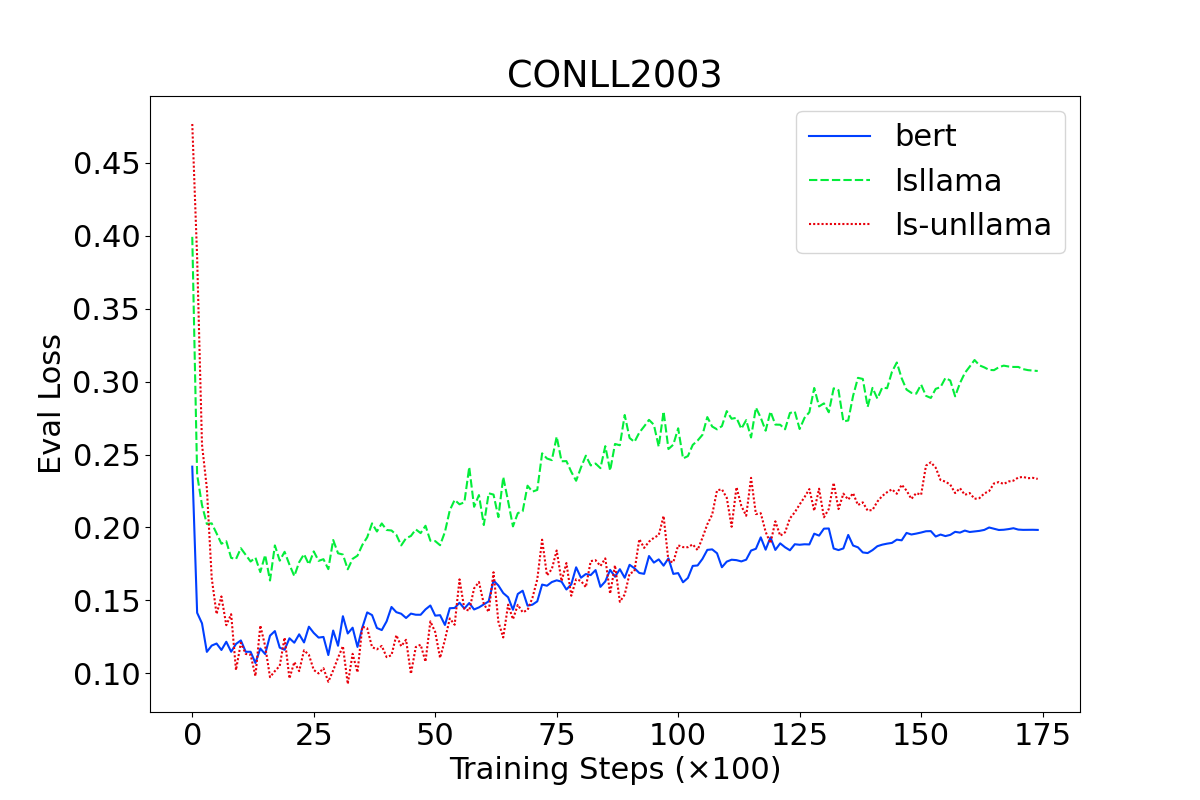}
     \end{subfigure}
        \caption{Testing loss of BERT, LS-LLaMA, and LS-unLLaMA in 10 training epochs.}
        \label{fig:sst5_conll_eval_loss}
\end{figure}

\begin{figure}[htbp]
     \centering
     \begin{subfigure}[b]{0.45\textwidth}
         \centering
         \includegraphics[width=\textwidth]{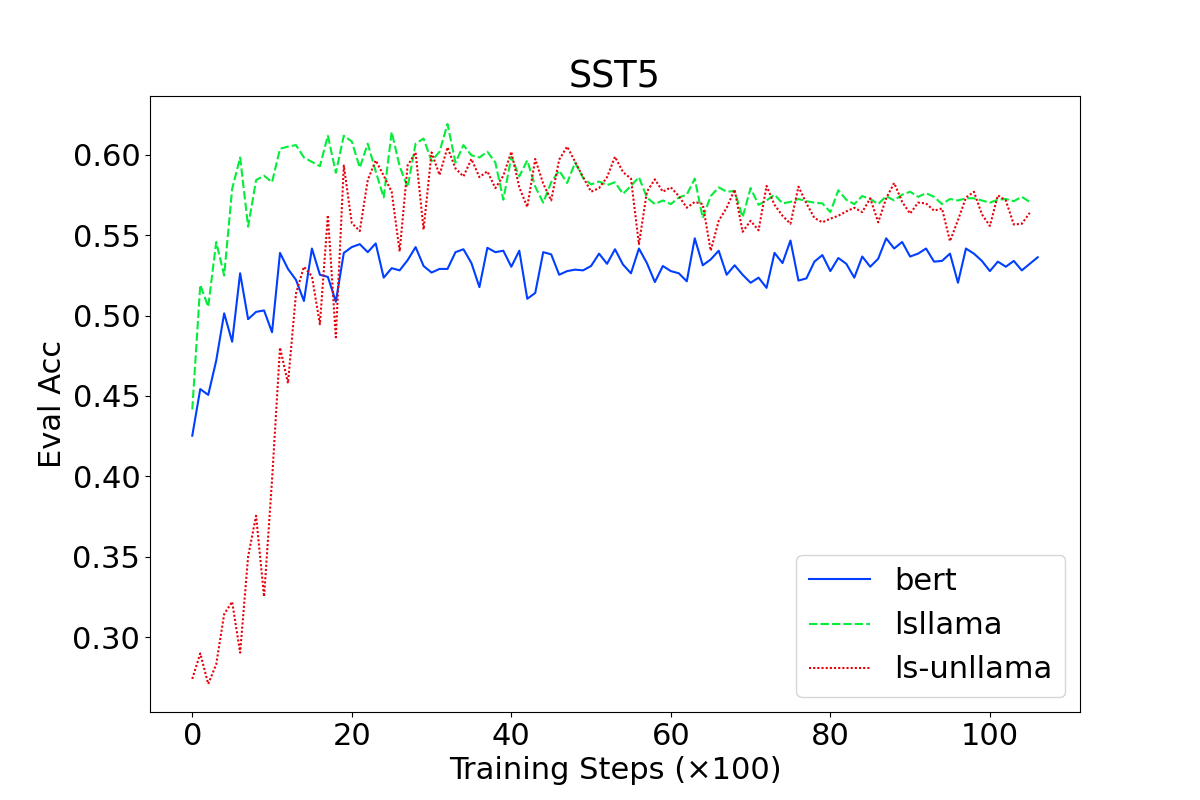}
     \end{subfigure}
     \begin{subfigure}[b]{0.45\textwidth}
         \centering
         \includegraphics[width=\textwidth]{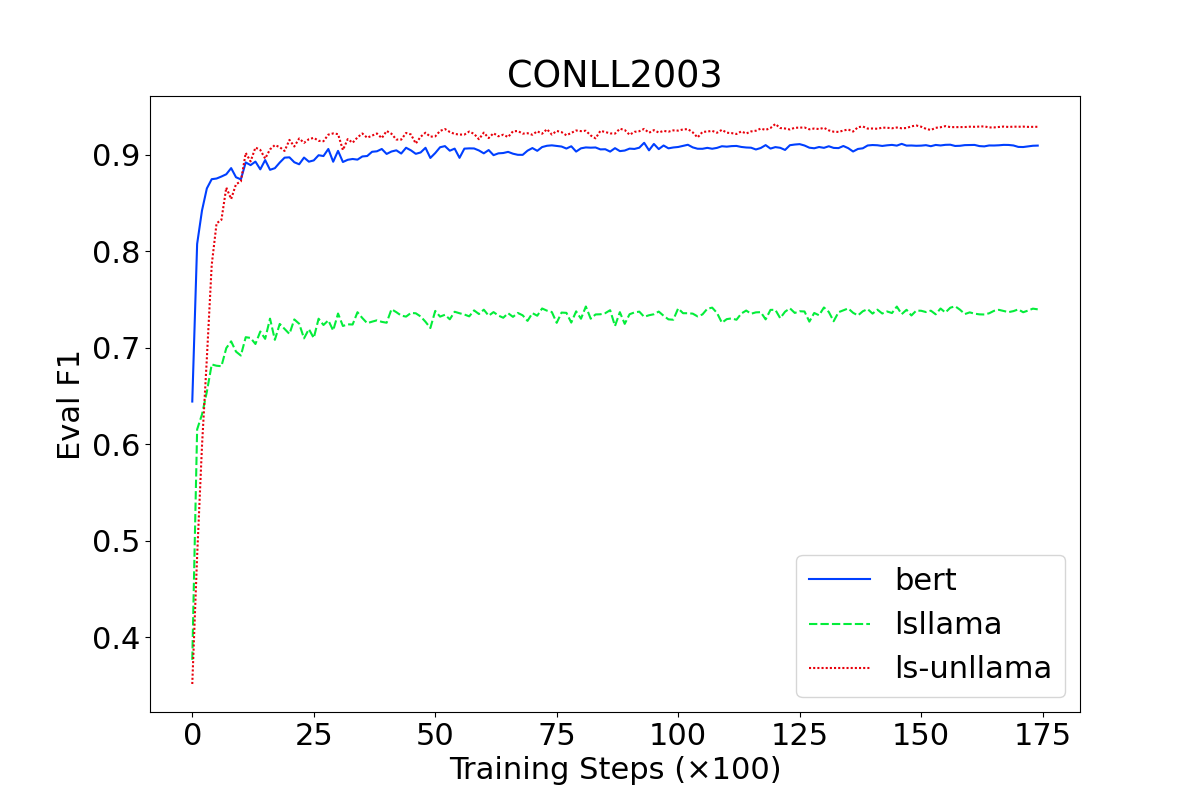}
     \end{subfigure}
        \caption{Testing accuracy / F1 score of BERT, LS-LLaMA, and LS-unLLaMA in 10 training epochs.}
        \label{fig:sst5_conll_eval_accu}
\end{figure}

We notice that the proposed approaches can outperform BERT on a range of benchmark datasets, hence it is interesting to observe how the LLaMA is finetuned in the same way as BERT. Therefore, we study the training process in 10 epochs of LS-LLaMA, LS-unLLaMA, and BERT on SST5 and ConLL2003 datasets. We recorded the training loss, evaluation loss, and evaluation accuracy / F1 score every 100 training steps and depict the data in Figures \ref{fig:sst5_conll_train_loss}, \ref{fig:sst5_conll_eval_loss}, and \ref{fig:sst5_conll_eval_accu}. On SST5, in which all these three models performs well, LS-LLaMA and BERT converge faster than LS-unLLaMA, and LS-unLLaMA's testing loss and accuracy tend to be unstable after 5,000 training steps. As discussed in Section \ref{sec:multiclass_result}, LS-unLLaMA underperforms on small-scale datasets like SST5, as it struggles to reconstruct the self-attention weights that were not well trained during pretraining because of the causal masks. Small datasets cannot provide sufficient label supervision and leads to severer overfitting problem. In contrast, the training process of LS-LLaMA is more stable, and it suffers less overfitting than BERT.

In NER task, the training process of LS-LLaMA is more unstable than those of LS-unLLaMA and BERT, as the token representations of LS-LLaMA intrinsically lacks global dependency due to the causal masks. We can also observe that the testing loss of LS-unLLaMA bounces back later than BERT, suggesting that LS-unLLaMA is more resistant to overfitting issue than BERT with a large training set. 

\subsubsection{Model size}

In light of the outstanding performance achieved by the smaller-scale 7B LLaMA-2, one may be curious about whether a larger LLM like 13B LLaMA-2 would yield even more impressive results. Therefore, we experimented the proposed label-supervised adaptation approach based on LLaMA-2-13B on sequence and token classification tasks. The results were listed in Tables \ref{tab:multiclass} and \ref{tab:ner}. 

In summary, the application of our label-supervised adaptation approach to 13B LLaMA-2, both with and without the causal mask, did not yield substantial improvements on either of the tasks. In the realm of sequence classification tasks, LS-LLaMA-2-13B does exhibit better performance compared to BERT and RoBERTa, as well as its 7B counterpart, on larger datasets like SST2 and AGNews. However, it falls short of the results achieved by LS-unLLaMA-2-7B. On smaller datasets such as Twitter-Fin and SST5, both 13B LS-LLaMA-2 and LS-unLLaMA fail to surpass their 7B counterparts. Moreover, LS-unLLaMA-2-13B also suffers from significant performance degradation, possibly caused by the insufficient volume of training set. 

The challenge is also evident in token classification tasks. Finetuning the 13B LLaMA model, employing cross-entropy loss and LoRA, leads to deteriorated performance than finetuning the 7B LLaMA and BERT in the same manner. These observations point to the challenge of fine-tuning the 13B LLaMA-2, particularly when the training samples are limited. The scarcity of training samples may cause severe overfitting problems on the limited training set. 

In essence, our experiments show that the performance of LLaMA-2 under label-supervised adaptation does not exhibit the expected linear scalability with an increase in model size. The availability of sufficient training samples becomes a critical determinant in applying label-supervised adaption for larger LLMs.

\section{Related Work}
Continuous efforts have been devoted to improving the problem-solving abilities of LLMs with their superior text generation capacity. One of the main research directions focuses on prompt engineering, which aims to generate higher-quality responses by harnessing LLMs' existing knowledge. Without modifying the parameters, one can give instructions with a few input-output exemplars or carefully crafted in-context prompts to help the model better understand the task and elicit a profound inferencing and reasoning process. Various prompting techniques have been proposed, such as contextual calibration \citep{zhao2021calibrate}, prompt programming \citep{reynolds2021prompt}, chat-based prompt \citep{wei2023zero}, chain-of-thought  \citep{wei2022chain}, and tree-of-thought \citep{yao2023tree}. These methods are especially preferred for extra-large LLMs like ChatGPT and PaLM-540B, on which tuning the parameters may be less feasible, they have been effective in a range of tasks including information extraction \citep{li2023evaluating}, semantic textual similarity \cite{li2023angle}, and reasoning \citep{kojima2022large}.

Despite the success of zero- and few-shot settings, LLMs frequently struggle in domains that require specific knowledge or precise response generation. To generalize LLMs on more downstream tasks, researchers have also investigated various instruction-tuning methods \citep{brown2020language,wei2021finetuned,wang2022self,peng2023instruction,phang2023hypertuning,zadouri2023pushing} to replenish domain knowledge and enhance LLMs performance. Instruction-tuning approaches tune the pretrained parameters with instructional data, which contain instructional commands and human-annotated expected outcomes \citep{sanh2021multitask,wang2022super}. \cite{peng2023instruction} finetuned LLaMA using 52K English and Chinese instruction-following instances generated using GPT-4. \cite{phang2023hypertuning} proposed HyperTuning that uses a hypermodel to generate task-specific parameters for a fixed downstream model for model adaptation. Instruction-following LLMs present substantial improvements in zero-shot performance on unseen tasks.

So far, no attempts have been made to finetune an LLM with discriminant labels. Our study verifies the feasibility of the proposed label-supervised adaptation approach on sequence and token classification tasks.

\section{Conclusion}

In conclusion, this paper embarks on a comprehensive exploration of label-supervised adaptation for enhancing LLMs' performance in both sequence and token classification, surpassing existing approaches like prompt engineering and instruction-tuning. Our two proposed variants, LS-LLaMA and LS-unLLaMA, have demonstrated remarkable and consistent improvements over robust benchmarks such as BERT and RoBERTa, across a range of text classification and under multilingual settings. With causal masks removed in the decoders, LS-unLLaMA yields state-of-the-art performance on token classification tasks like NER. This study depicts the potential of LLMs as robust text encoders, with latent representations can be applied in a broader spectrum of applications when explicit label supervision is provided.

The implications of our findings may extend beyond the specific tasks explored in this study. The proposed label-supervised adaptation offers an accessible and highly effective configuration that can serve as a novel interface for various downstream tasks, such as domain-specific text classification and token classification, by finetuning a small-scale LLMs such as LLaMA-2-7B. This method could potentially reshape the landscape of LLM applications.

\bibliographystyle{unsrtnat}
\bibliography{references}  






\end{document}